\documentclass[letterpaper, 10 pt, conference]{ieeeconf}  

\IEEEoverridecommandlockouts                              

\usepackage{import}
\usepackage{cite}
\usepackage{graphics} 
\usepackage{epsfig} 
\usepackage{times} 
\usepackage{amsmath} 
\usepackage{amssymb}  
\usepackage{lipsum}
\usepackage{multicol}
\usepackage{graphicx}
\usepackage{afterpage}
\usepackage{subcaption}
\usepackage{multirow}
\usepackage{layout}
\usepackage{gensymb}
\usepackage{graphicx}
\usepackage[dvipsnames]{xcolor}
\usepackage{amsmath}
\usepackage{varwidth}
\usepackage{tikz}
\usetikzlibrary{shapes,arrows,positioning}
\usepackage{booktabs}

\usepackage{siunitx}
\usepackage[nolist]{acronym}
\newcommand{\suma}{$+$}
\newcommand{\diffa}{$-$}

\newcommand{\down}{$\displaystyle \downarrow$}
\newcommand{\mux}{}

\tikzstyle{every node}=[font=\scriptsize]
\tikzset{%
	line/.style = {draw, thick},
  block/.style    = {draw, thick, rectangle, minimum height = 1em, 
    minimum width = 3em},
  block3/.style    = {draw, thick, rectangle, minimum height = 3.5em, 
    minimum width = 3em},
  mux/.style     = {draw, rectangle, fill, minimum height = 8.5em, minimum width = 0.5em},
  mux2/.style     = {draw, rectangle, fill, minimum height = 2em, minimum width = 0.5em},
  mux3/.style     = {draw, rectangle, fill, minimum height = 5em, minimum width = 0.5em},
  sum/.style      = {draw, circle}, 
  diff/.style      = {draw, circle}, 
  down/.style      = {draw, thick, rectangle}, 
  branch/.style = {coordinate}, 
  branchpt/.style = {coordinate, circle, fill, minimum size=3pt,inner sep=0pt}
}

\DeclareMathOperator*{\argmin}{argmin}   

\usepackage{fancyhdr}
\fancypagestyle{pageStyleOne}{%
    \fancyhf{}
}


\title{\LARGE \bf
Vision-Aided Absolute Trajectory Estimation Using an Unsupervised Deep Network with Online Error Correction
}

\author{E. Jared Shamwell$^{1}$, Sarah Leung$^{2}$, William D. Nothwang$^{3}$
\thanks{*This work was supported by the US Army Research Laboratory}
\thanks{$^{1}$E. Jared Shamwell, PhD is a research scientist with GTS stationed at the US Army Research Laboratory, Adelphi, MD 20783.
        {\tt\small earl.j.shamwell.ctr@mail.mil; ejsham@umd.edu}}%
\thanks{$^{2}$Sarah Leung is a research scientist with GTS stationed at the US Army Research Laboratory, Adelphi, MD 20783.
        {\tt\small sarah.leung.ctr@mail.mil}}%
\thanks{$^{3}$William D. Nothwang, PhD is the Branch Chief (a) of the Micro and Nano Devices and Materials Branch in the Sensors and Electron Devices Directorate at the US Army Research Laboratory, Adelphi, MD 20783.
        {\tt\small william.d.nothwang.civ@mail.mil }}%
}

\begin{document}

\maketitle
\thispagestyle{pageStyleOne} 
\pagestyle{fancy} 

\begin{acronym}[EMBC]
	\acro{ARL}{Army Research Laboratory}
	\acro{IMU}{inertial measurement unit}
	\acroplural{IMU}[IMUs]{inertial measurement units}
	\acro{SOA}{state-of-the-art}
	\acro{VSLAM}{visual simultaneous localization and mapping}
	\acro{VO}{visual odometry}
	\acro{VIO}{visual-inertial odometry}
	\acro{RTK}{Real Time Kinematic}
	\acro{SLAM}{Simultaneous localization and mapping}
	\acro{EKF}{Extended Kalman Filter}
	\acro{MSCKF}{Multi-state Constraint Kalman Filter}
	\acro{MDN}{mixture density network}
	\acro{CNN}{convolutional neural network}
	\acro{WTA}{winner-take-all}
	\acro{RMSE}{root mean squared error}
	\acro{CNN}{convolutional neural network}
\end{acronym}

\begin{abstract}

We present an unsupervised deep neural network approach to the fusion of RGB-D imagery with inertial measurements for absolute trajectory estimation. Our network, dubbed the Visual-Inertial-Odometry Learner (VIOLearner), learns to perform \acf{VIO} without \acf{IMU} intrinsic parameters (corresponding to gyroscope and accelerometer bias or white noise) or the extrinsic calibration between an \ac{IMU} and camera. The network learns to integrate \ac{IMU} measurements and generate hypothesis trajectories which are then corrected online according to the Jacobians of scaled image projection errors with respect to a spatial grid of pixel coordinates. We evaluate our network against \acf{SOA} visual-inertial odometry, visual odometry, and \acf{VSLAM} approaches on the KITTI Odometry dataset \cite{Geiger2013} and demonstrate competitive odometry performance.

\end{abstract}

\section{Introduction}

Originally coined to characterize honey bee flight \cite{srinivasan1991}, the term ``visual odometry'' describes the integration of apparent image motions for dead-reckoning-based navigation (literally, as an odometer for vision). While work in what came to be known as \acf{VO} began in the 1980s for the Mars Rover project, the term was not popularized in the engineering context until around 2004 \cite{nister2004}. 

Approaches in \ac{VIO} combine visual estimates of motion with those measured by multi-axis accelerometers and gyroscopes in \acfp{IMU}. As \acp{IMU} measure only linear accelerations and angular velocities, inertial approaches to localization are prone to exponential drift over time due to the double integration of accelerations into pose estimates. Combining inertial estimates of pose change with visual estimates allows for the `squashing' of inertial estimates of drift.

\begin{figure}
\input{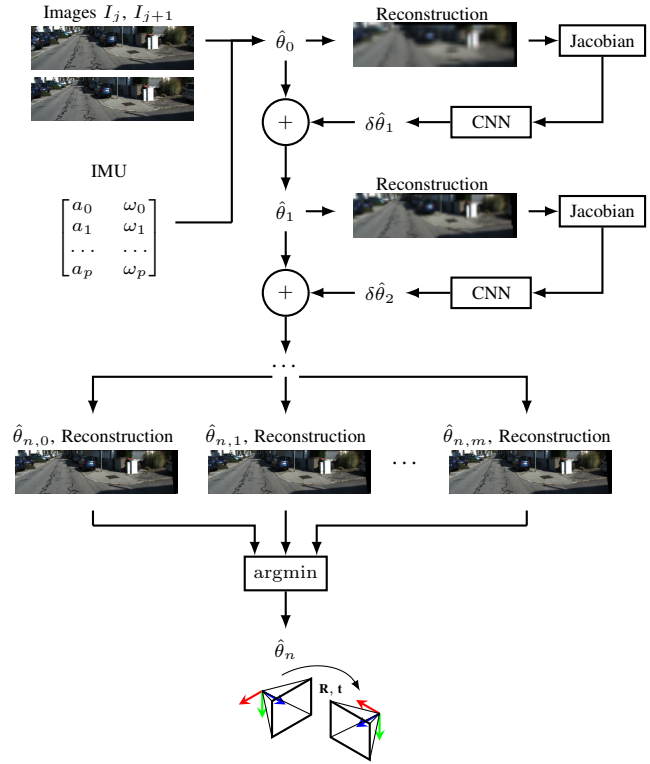}

\caption{Generalized overview of our VIOLearner network (see ÒFig. \ref{fig:voi_diagram}Ò for a more detailed/accurate representation of the architecture). Note the hierarchical architecture where $\theta$, the true change in pose between images, is estimated at multiple scales and then corrected via convolutional processing of the Jacobian of Euclidean projection error at that scale with respect to a grid of source coordinates. The final Level $n$ computes $m$ hypothesis reconstructions (and associated estimates $\hat{\theta}_{n,m}$) and the $\hat{\theta}_{n,m}$ with the lowest paired Euclidean projection error between the reconstruction and true target image $I_{j+1}$ is output as the final network estimate $\hat{\theta}_{n}$ .}
\end{figure}

While \ac{VO}, \ac{VIO}, and \acf{VSLAM} can be performed by a monocular camera system, the scale of trajectory estimates cannot be directly estimated as depth is unobservable from monocular cameras (although for \ac{VIO}, integrating raw \acp{IMU} measurements can provide noisy, short-term estimates of scale). Generally, depth can be estimated from RGB-D cameras or from stereo camera systems. We have chosen to include depth in our input domain as a means of achieving absolute scale recovery: while absolute depth can be generated using an onboard sensor, the same cannot be said for the change in pose between image pairs. 

GPS, proposed as a training pose-difference signal in \cite{pillai2017}, has limited accuracy on the order of meters which then inherently limits the accuracy of a GPS-derived inter-image training signal for bootstrapping purposes. While \acf{RTK} GPS can reduce this error to the order of centimeters, \ac{RTK} GPS solutions require specialized localized base-stations. Scene depth, on the other hand, can be accurately measured  by an RGB-D sensor (e.g. Microsoft Kinect, Asus Xtion, or new indoor/outdoor Intel RealSense cameras), stereo cameras, or LIDAR (e.g. a Velodyne VLP-16 or PUCK Lite) directly onboard a vehicle.

The main contribution of this paper is the unsupervised learning of trajectory estimates with absolute scale recovery from RGB-D + inertial measurements with

\begin{itemize}
\item Built-in online error correction modules;
\item Unknown \ac{IMU}-camera extrinsics; and
\item Loosely temporally synchronized camera and \ac{IMU}.
\end{itemize}

\begin{figure*}[t]
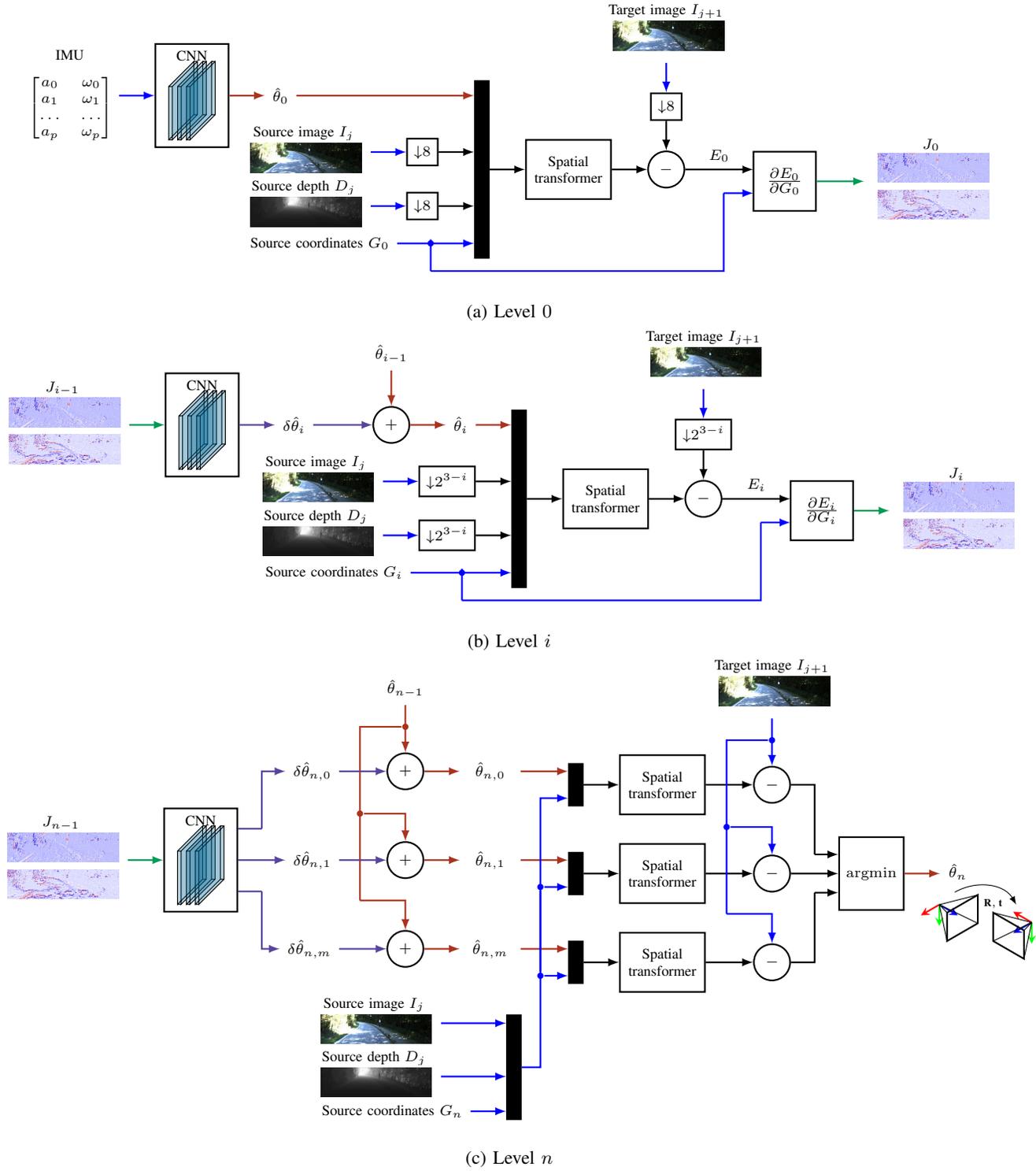

\centering
\begin{subfigure}[t]{1\textwidth}
	\input{tex_figures/layer_0.tikz}	
   	\caption{Level $0$}
   	\label{fig:net_0} 
\end{subfigure}
\begin{subfigure}[t]{1\textwidth}
	\input{tex_figures/layer_i.tikz}	
   	\caption{Level $i$ }
	\label{fig:net_i}
\end{subfigure}
\begin{subfigure}[t]{1\textwidth}
	\input{tex_figures/layer_n.tikz}		
   	\caption{Level $n$}
   	\label{fig:net_n}
\end{subfigure}

\caption{VIOLearner: (a) \ac{IMU} data is fed into a series of \acf{CNN} layers, which output a 3D affine matrix estimate of change in camera pose $\hat{\theta}$ between a source image $I_{j}$ and target image $I_{j+1}$. The transform $\hat{\theta}$ is applied to a downsampled source image via a spatial transformer module and the Euclidean error between the spatial transformation and the downsampled target image is computed as $E_{0}$. The Jacobian $\frac{\partial E_{0}}{\partial G_{0}}$ of the error image $E_{0}$ is taken with respect to the source coordinates $G_{0}$, and fed to the next level. (b) The Jacobian $\frac{\partial E_{i-1}}{\partial G_{i-1}}$ from the previous level is fed through \ac{CNN} layers to produce an additive refinement $\partial{\hat{\theta}_{i}}$ that is summed with the previous transform estimate $\hat{\theta}_{i-1}$ to produce a new transform estimate $\hat{\theta_{i}}$. The Jacobian $\frac{\partial E_{i}}{\partial G_{i}}$ is propagated forward. (c) In the final level, the previous Jacobian $\frac{\partial E_{n-1}}{\partial G_{n-1}}$ is processed through \ac{CNN} layers for $m$ hypothesis refinements.}

\label{fig:voi_diagram}
\end{figure*}

\section{Related Work} \label{related_work}

\subsection{Traditional Methods}
In \ac{VO} and \ac{VSLAM}, only data from camera sensors is used and tracked across frames to determine the change in the camera pose. \acf{SLAM} approaches typically consist of a front end, in which features are detected in the image, and a back end, in which features are tracked across frames and matched to keyframes to estimate camera pose, with some approaches performing loop closure as well. ORB-SLAM2 \cite{mur2017} is a visual \ac{SLAM} system for monocular, stereo, and RGB-D cameras. It used bundle adjustment and a sparse map for accurate, real-time performance on CPUs. ORB-SLAM2 performed loop closure to correct for the accumulated error in its pose estimation. ORB-SLAM2 has shown \acf{SOA} performance on a variety of \acf{VO} benchmarks. 

In \ac{VIO} and visual-inertial \ac{SLAM}, the fusion of imagery and \ac{IMU} measurements are typically accomplished by filter-based approaches or non-linear optimization approaches. ROVIO \cite{bloesch2017} is a \ac{VIO} algorithm for monocular cameras, using a robust and efficient robocentric approach in which 3D landmark positions are estimated relative to camera pose. It used an \acf{EKF} to fuse the sensor data, utilizing the intensity errors in the update step. However, because ROVIO is a monocular approach, accurate scale is not recovered. OKVIS \cite{leutenegger2015} is a keyframe-based visual-inertial \ac{SLAM} approach for monocular and stereo cameras. OKVIS relied on keyframes, which consisted of an image and estimated camera pose, a batch non-linear optimization on saved keyframes, and a local map of landmarks to estimate camera egomotion. However, it did not include loop closure, unlike the \ac{SLAM} algorithms with \ac{SOA} performance. 

There are also several approaches which enhance \ac{VIO} with depth sensing or laser scanning. Two methods of depth-enhanced \ac{VIO} are built upon the \ac{MSCKF} \cite{mourikis2007, li2013ekf} algorithm for vision-aided inertial navigation, which is another \ac{EKF}-based \ac{VIO} algorithm. One method is the \ac{MSCKF}-3D \cite{galfond2014} algorithm, which used a monocular camera with a depth sensor, or RGB-D camera system. The algorithm performed online time offset correction between camera and \ac{IMU}, which is critical for its estimation process, and used a Gaussian mixture model for depth uncertainty. Pang et al. \cite{pang2017} also demonstrated a depth-enhanced \ac{VIO} approach based on \ac{MSCKF}, with 3D landmark positions augmented with sparse depth information kept in a registered depth map. Both approaches showed improved accuracy over \ac{VIO}-only approaches. Finally, Zhang and Singh \cite{zhang2017} proposed a method for leveraging data from a 3D laser. The approach utilized a multi-layer pipeline to solve for coarse to fine motion by using \ac{VIO} as a subsystem and matching scans to a local map. It demonstrated high position accuracy and was robust to individual sensor failures. 

\subsection{Learning-Based Methods}
Recently, there have been several successful unsupervised approaches to depth estimation that are trained using reconstructive loss \cite{garg2016, godard2017} from image warping similar to our own network. Garg et. al and Godard et. al used stereo image pairs with known baselines and reconstructive loss for training. Thus, while technically unsupervised, the known baseline effectively provides a known transform between two images. Our network approaches the same problem from the opposite direction: we assume known depth and estimate an unknown pose difference. 

Pillai and Leonard \cite{pillai2017} demonstrated visual egomotion learning by mapping optical flow vectors to egomotion estimates via a \ac{MDN}. Their approach not only required optical flow to already be externally generated (which can be very computationally expensive), but also was trained in a supervised manner and thus required the ground truth pose differences for each exemplar in the training set. 

SFMLearner \cite{zhou2017} demonstrated the unsupervised learning of unscaled egomotion and depth from RGB imagery. They input a consecutive sequence of images and output a change in pose between the middle image of the sequence and every other image in the sequence, and the estimated depth of the middle image. However, their approach was unable to recover the scale for the depth estimates or, most crucially, the scale of the changes in pose. Thus, their network's trajectory estimates needed to be scaled by parameters estimated from the ground truth trajectories and in the real-world, this information will of course not be available. SFMLearner also required a sequence of images to compute a trajectory. Their best results were on an input sequence of five images whereas our network only requires a source-target image pairing.

UnDeepVO \cite{li2017} is another unsupervised approach to depth and egomotion estimation. It differs from \cite{zhou2017} in that it was able to generate properly scaled trajectory estimates. However, unlike \cite{zhou2017} and similar to \cite{garg2016, godard2017}, it used stereo image pairs for training where the baseline between images is known and thus, UnDeepVO can only be trained on datasets where stereo image pairs are available. Additionally, the network architecture of UnDeepVO cannot be extended to include motion estimates derived from inertial measurements because the spatial transformation between paired images from stereo cameras are unobservable by an \ac{IMU} (stereo images are recorded simultaneously).

VINet \cite{clark2017} was the first end-to-end trainable visual-inertial deep network. While VINet showed robustness to temporal and spatial misalignments of an \ac{IMU} and camera, it still required extrinsic calibration parameters between camera and \ac{IMU}. This is in contrast to our VIOLearner which requires no \ac{IMU} intrinsics or \ac{IMU}-camera extrinsics. In addition, VINet was trained in a supervised manner and thus required the ground truth pose differences for each exemplar in the training set which are not always readily available.

\section{Approach} \label{approach}

VIOLearner is an unsupervised \ac{VIO} deep network that estimates the scaled egomotion of a moving camera between some time $t_{j}$ at which source image $I_{j}$ is captured and time $t_{j+1}$ when target image $I_{j+1}$ is captured. VIOLearner receives an input RGB-D source image, a target RGB image, \ac{IMU} data from $t_{j-1}$ to $t_{j+1}$, and a camera calibration matrix $K$ with the camera's intrinsic parameters. With access to $K$, VIOLearner can generate camera pose changes in the camera frame using a view-synthesis approach where the basis for its training is in the Euclidean loss between a target image and a reconstructed target image generated using pixels in a source image sampled at locations determined by a learned 3D affine transformation (via a spatial transformer of \cite{Jaderberg2015}). 

\subsection{Multi-Scale Projections and Online Error Correction} \label{mult_scale}

Similar to \cite{zhou2017}, VIOLearner performs multi-scale projections. We scale projections by factors of $8$, $4$, and $2$. However, in our network, multi-scale projections not only help to overcome gradient locality during training (see \cite{zhou2017, anandan1993} for a broader discussion of this well-known issue), but also aid in online error correction at runtime.

Generally, at each level, the network computes the Jacobians of the reprojection error at that level with respect to the grid of coordinates. Convolutions are then performed on this Jacobian (sized HxWx$2$) and a $\delta \hat{\theta}$ is computed. This $\delta \hat{\theta}$ is added to the previously generated affine matrix $\hat{\theta}$. This is repeated for each level in the hierarchy. VIOLearner uses a total of $5$ levels and additional detail on the computations performed at each level is provided in the following sections.


\subsection{Level 0}
VIOLearner first takes raw \ac{IMU} measurements and learns to compute an estimated 3D affine transformation $\hat{\theta}_0$ that will transform a source image $I_{j}$ into a target image $I_{j+1}$ (the top left of ÒFig. \ref{fig:net_0}Ò). The network downsamples (\down{} in ÒFig. \ref{fig:voi_diagram}Ò) the source image $I_{j}$ (and associated depth and adjusted camera matrix $K$ for the current downsampling factor) by a factor of 8 and applies the 3D affine transformation to a normalized grid of source coordinates $[X_{src},Y_{src}]$ to generate a transformed grid of target coordinates $[X_{tgt},Y_{tgt}]$.  

VIOLearner then performs bilinear sampling to produce a reconstruction image $Ir$ by sampling from the source image $I_{j}$ at coordinates $(x,y) \in [X_{tgt},Y_{tgt}]$:

\begin{equation}
Ir(x,y) =  \sum_{h,w}^{H, W} I_{j}(h,w) \phi(x-h, y-w) 
\end{equation}

\noindent where the function $\phi$ is zero except in the range where it is non-negative and equal to one at $ (0,0) $.

As in \cite{Jaderberg2015}, by only evaluating at the sampling pixels, we can therefore compute sub-gradients that allow error back-propagation to the affine parameters $\hat{\theta}_0$ by computing error with respect to the coordinate locations $(x,y)$ instead of the more conventional error with respect to the pixel intensity:

\begin{equation} \label{dx}
  \frac{\partial Ir^c_j}{\partial x^s_{j}}= \\
  \sum_{h,w}^{H,W} U(h,w) \max(0, 1-|y-w|)  \begin{cases}
    1, & \text{if $h>=x$}.\\
    -1, & \text{if $h<x$}.
  \end{cases} 
\end{equation}

\begin{equation} \label{dy}
  \frac{\partial Ir^c_j}{\partial y^s_{j}}= \\
  \sum_{h,w}^{H,W} U(h,w) \max(0, 1-|x-h|)  \begin{cases}
    1, & \text{if $w>=y$}.\\
    -1, & \text{if $w<y$}.
  \end{cases} 
\end{equation}

Starting with the Level 0, the Euclidean loss is taken between the downsampled reconstructed target image and the actual target image. For Level 0, this error is computed as:

\begin{equation} \label{level0_error}
E_{0} = \lVert I_{r}^{0} - I_{j+1}^{0} \rVert ^ 2 
\end{equation}

The final computation performed by Level 0 is of the Jacobian of the Euclidean loss of ÒEquation \ref{level0_error}Ò with respect to the source coordinates $G_{0}$ from ÒEquation \ref{dx}Ò and ÒEquation \ref{dy}Ò. The resulting Jacobian matrix $J_{0}$ has the same dimensionality of the grid of source coordinates $G_{0}$ ($\frac{H}{8}$x$\frac{W}{8}$x$2$) and is depicted in ÒFig. \ref{fig:net_0}Ò. In traditional approaches, the gradient and error equations above are only used during training. However, VIOLearner is novel in that it also computes and employs these gradients during each inference step of the network. During both training and inference, the Jacobian $J_{0}$ is computed and passed to the next level for processing.

\subsection{Levels i to n-1}
For the second level in the network through the n-$1$ level, the previous level's Jacobian $J_{i-1}$ is input and processed through layers of convolutions to generate a $\partial \hat{\theta}_{i}$. This $\partial \hat{\theta}_{i}$ represents a computed correction to be applied to the previous level's 3D affine transform $\hat{\theta}_{i-1}$. $\partial \hat{\theta}_{i}$ is summed with $\theta_{i-1}$ to generate $\hat{\theta}_{i}$ which is then applied to generate a reconstruction that is downsampled by a factor $2^{3-i}$. Error is again computed as above in ÒEquation \ref{level0_error}Ò and the Jacobian is similarly found as it was in Level 0 and input to the next Level $i+1$.

\subsection{Level n and Multi-Hypothesis Pathways} \label{multi_hypothesis}

The final level of VIOLearner employs multi-hypothesis pathways similar to \cite{shamwell2017b, shamwell2018a} where several possible hypotheses for the reconstructions of a target image (and the associated transformations $\hat{\theta}_{m}, m \in M$ which generated those reconstructions) are computed in parallel. The lowest error hypothesis reconstruction is chosen during each network run and the corresponding affine matrix $\hat{\theta}_{m*}$ which generated the winning reconstruction is output as the final network estimate of camera pose change between images $I_{j}$ and $I_{j+1}$.


This multi-hypothesis approach allows the network to generate several different pathways and effectively sample from an unknown noise distribution. For example, as \acp{IMU} only measure linear accelerations, they fail to accurately convey motion during periods of constant velocity. Thus, a window of integrated \ac{IMU} measurements are contaminated with noise related to the velocity at the beginning of the window. With a multi-hypothesis approach, the network has a mechanism to model uncertainty in the initial velocity (see ÒSection \ref{multi_hypoth_disc}Ò for a discussion).

Error for this last multi-hypothesis level is computed according to a \ac{WTA} Euclidean loss rule (see \cite{shamwell2017b} for more detail and justifications):

\begin{equation}
I_{r}^{n*} \longleftarrow \argmin_k \lVert I_{r}^{n,k} - I_{j+1}^{n} \rVert ^ 2
\end{equation}
\begin{equation}
E_{n} = \lVert I_{r}^{n,*} - I_{j+1}^{n} \rVert ^ 2 
\end{equation}

\noindent where $I_{r}^{n,*}$ is the lowest error hypothesis reconstruction. Loss is then only computed for this one hypothesis pathway and error is backpropagated only to parameters in that one pathway. Thus, only parameters that contributed to the winning hypothesis are updated and the remaining parameters are left untouched.

The final loss $\mathcal{L}$ by which the network is trained is then simply the sum of the Euclidean loss terms for each level plus a weighted L1 penalty over the bias terms which we empirically found to better facilitate training and gradient back-propagation:

\begin{equation}
\mathcal{L} = \sum_{i=0}^{n} E_{i} + \lambda * \lvert bias \rvert
\end{equation}

\section{Methods} \label{methods}


\subsection{Network Architecture}

\subsubsection{\ac{IMU} Processing}
The initial level of VIOLearner uses two parallel pathways of $7$ convolutional layers for the \ac{IMU} angular velocity and linear accelerations, respectively. Each pathway begins with $2$ convolutional layers each of $64$ single-stride $3$x$5$ filters on the $batch$x$20$x$3$ \ac{IMU} angular velocity or linear accelerations followed by $2$ convolutional layers of $128$ filters each of stride $2$ with the same $3$x$5$ kernel. Next, $3$ convolutional layers of $256$ filters are applied with strides of $2$, $1$, and $1$, and kernels of size $3$x$5$, $3$x$3$, and $3$x$1$. The final convolutional layer in the angular velocity and linear acceleration pathways were flattened into $batch$x$1$x$3$ tensors using a convolutional layer with three filters of kernel size $1$ and stride $1$ before being concatenated together into a tensor $pose\_imu$. 

\subsubsection{3D Affine Transformations}
The first three components in \textit{pose\_imu} correspond to rotations $[\alpha_{0}, \beta_{0}, \gamma_{0}]$ representing rotations about the x, y, and z axis respectively. Rotation matrices are computed as

\begin{equation} \label{rz}
R_{x}^{0}(\alpha_{0}) =
\left[\begin{array}{rrr}
1 & 0 & 0 \\
0 & \cos{\alpha_{0}} & -\sin{\alpha_{0}} \\
0 & \sin{\alpha_{0}} & \cos{\alpha_{0}} \\
\end{array} \right]
\end{equation}

\begin{equation} \label{ry}
R_{y}^{0}(\beta_{0}) =
\left[\begin{array}{rrr}
\cos{\beta_{0}} & 0 & \sin{\beta_{0}} \\
0 & 1 & 0 \\
-\sin{\beta_{0}} & 0 & \cos{\beta_{0}} \\
\end{array} \right]
\end{equation}

\begin{equation} \label{rz}
R_{z}^{0}(\gamma_{0}) =
\left[\begin{array}{rrr}
\cos{\gamma_{0}} & -\sin{\gamma_{0}} & 0 \\
\sin{\gamma_{0}} & \cos{\gamma_{0}} & 0 \\
0 & 0 & 1 \\
\end{array} \right]
\end{equation}

\noindent and a 3D rotation matrix is generated as

\begin{equation} \label{r0}
R^{0} = R_{z}*R_{y}*R_{x} 
\end{equation}


The last three elements in $pose\_imu$ directly correspond to a translation vector $T^{0} = [x_{0},y_{0},z_{0}]^\top$. Together with the 3D rotation matrix $R^{0}$, we finally form a 4x4 homogeneous transformation matrix $\hat{\theta}_0$ as

\begin{equation}
\hat{\theta}_0 = \left[
\begin{array}{cc}
R^{0} & T^{0} \\ 
0 & 1 \\
\end{array}
\right]
\end{equation}
\subsubsection{Online Error Correction and Pose Refinement}
For each $\frac{H}{s}$x$\frac{W}{s}$x$2$ Jacobian matrix $\frac{\partial{E_{i}}}{\partial{G_{i}}}$ at all scales $s \in S$, three convolutional layers of $128$ filters are applied with kernel sizes $7$x$7$, $5$x$5$, and $3$x$3$ and strides of $1$, $2$, and $2$, respectively. Then, an additional $2$ to $5$ convolutional layers are applied depending on the downsampling factor of the current level with the number of additional layers increasing as the downsampling factor decreases (the final level using $5$ additional convolutional layers). For each level, the final layer generates a pose estimate $pose\_refinement_{i}$ using a single-strided convolution with a kernel size of $1$.

The outputs $pose\_refinement_{i}$ are split similarly to $pose\_imu$ into rotations $[\partial\alpha_{i}, \partial\beta_{i}, \partial\gamma_{i}]$ and translation $T^{i} = [\partial{x_{i}},\partial{y_{i}},\partial{z_{i}}]^\top$.  The new rotations and translations for level $i$ are then computed as 
\begin{equation}
[\alpha_{i}, \beta_{i}, \gamma_{i}] = [\alpha_{i-1}, \beta_{i-1}, \gamma_{i-1}] + [\partial\alpha_{i}, \partial\beta_{i}, \partial\gamma_{i}]
\end{equation}

\begin{equation}
[x_{i},y_{i},z_{i}]^\top = [x_{i-1},y_{i-1},z_{i-1}]^\top + [\partial{x_{i}},\partial{y_{i}},\partial{z_{i}}]^\top
\end{equation}

This repeats for each level until the final level where $\hat{\theta_{n}}$ is output as the final estimate of the change in camera pose.

\subsubsection{Multi-Hypothesis Generation}
The final level of VIOLearner uses $4$ hypothesis pathways as described above in ÒSection \ref{multi_hypothesis}Ò.

\subsection{Training Procedures}

VIOLearner was trained for $100,000$ iterations (approximately $173$ epochs) using a batch size of 32. As the network was trained, we calculated error on the validation set at intervals of 500 iterations. The results presented in this paper are from a network model that was trained for $62,000$ iterations as it provided the highest performance on the validation set. We used the Adam solver with momentum1=$0.9$, momentum2=$0.99$, gamma=$0.5$, learning rate=$2\mathrm{e}{-4}$, and an exponential learning rate policy. The network was trained on a desktop computer with a 3.00 GHz Intel i7-6950X processor and Nvidia Titan X GPUs.

\subsection{KITTI Odometry Dataset}

We evaluate VIOLearner on the KITTI Odometry dataset \cite{Geiger2013} and used sequences $00-08$ for training excluding sequence $03$ because the corresponding raw file $2011\_09\_26\_drive\_0067$ was not online at the time of publication. Sequences 09 and 10 were withheld for the test set as was done in \cite{zhou2017}. Additionally, $5\%$ of KITTI sequences $00-08$ was withheld as a validation set. This left a total of $18,422$ training images, $2,791$ testing images, and $969$ validation images.

Depth is not available for the full resolution of KITTI images so we cropped each image in KITTI from $376$x$1241$ to $224$x$1241$ (and first resized each image to $376$x$1241$ if the resolution was different as is the case for certain sequences) and then scaled the cropped images to size $128$x$480$. In all experiments, we randomly selected an image for the source and used the successive image for the target. Corresponding \SI{100}{\hertz} \ac{IMU} data was collected from the KITTI raw datasets and for each source image, the preceding \SI{100}{\milli\second} and the following \SI{100}{\milli\second} of \ac{IMU} data was combined yielding a length $20$x$6$ vector (\SI{100}{\milli\second} prior to the source image and the approximately \SI{100}{\milli\second} between source and target image). We chose to include \ac{IMU} data in this way so that the network could learn how to implicitly estimate a temporal offset between camera and \ac{IMU} data as well as glean an estimate of the initial velocity $V_{0}$ at the time of source image capture by looking to previous data.

\section{Evaluation} \label{evaluation}


In the literature, there is no readily comparable approach with the exact input and output domains as our network (namely RGB-D + inertial inputs and scaled output odometry; see ÒSection \ref{related_work}Ò for approaches with our input domain but that do not publicly provide their code or evaluation results on KITTI). Nonetheless, we compare our approach to the following recent \ac{VO}, \ac{VIO}, and \ac{SLAM} approaches described earlier in ÒSection \ref{related_work}Ò:

\begin{itemize}
\item SFMLearner \cite{zhou2017}
\item VINet \cite{clark2017}
\item UnDeepVO \cite{li2017}
\item OKVIS \cite{leutenegger2015}
\item ROVIO \cite{bloesch2017}
\item ORB-SLAM2 \cite{mur2017}
\item VISO2-M (results reproduced from \cite{li2017})
\item ORB-SLAM-M (results reproduced from \cite{li2017})
\item EKF+VISO2 (results reproduced from \cite{clark2017})
\end{itemize}


For OKVIS, ROVIO, and SFMLearner, we temporally align each output and ground truth trajectory by cross correlating the norm of the rotational accelerations and perform 6-DOF alignment for the first tenth of the trajectory to ground truth. For ROVIO and SFMLearner, we also estimate the scale from ground truth.

\def\arraystretch{1.05}

\begin{table*}[t]
\caption{Comparisons to \ac{VO} approaches on KITTI sequences 00, 02, 05, 07, and 08. Results reproduced from \cite{li2017}. $t_{rel} (\SI{}{\percent})$ is the average translational error percentage on lengths \SI{100}{\meter} - \SI{800}{\meter} and $r_{rel} (\SI{}{\degree})$ is the rotational error $(\SI{}{\degree}/\SI{100}{\meter})$ on lengths \SI{100}{\meter} - \SI{800}{\meter}.  \cite{li2017} only reported results on KITTI Odometry sequences 00, 02, 05, 07, and 08 so in this table so we only report identical results for VIOLearner. }
\label{VO_results}
\begin{center}
\begin{tabular}{l*{12}{c}}
\hline
 & \multicolumn{2}{c}{\textit{VIOLearner}} & \multicolumn{2}{c}{\textit{UnDeepVO}} & \multicolumn{2}{c}{\textit{SFMLearner}} & \multicolumn{2}{c}{\textit{VISO2-M}} & \multicolumn{2}{c}{\textit{ORB-SLAM-M}} & \multicolumn{2}{c}{\textit{ORB-SLAM2}}\\
\cmidrule(lr){2-3} 
\cmidrule(lr){4-5} 
\cmidrule(lr){6-7} 
\cmidrule(lr){8-9} 
\cmidrule(lr){10-11} 
\cmidrule(lr){12-13} 

Seq\\
\hline
& $t_{rel} (\SI{}{\percent})$ & $r_{rel} (\SI{}{\degree})$ & $t_{rel} (\SI{}{\percent})$ & $r_{rel} (\SI{}{\degree})$ & $t_{rel} (\SI{}{\percent})$ & $r_{rel} (\SI{}{\degree})$ & $t_{rel} (\SI{}{\percent})$ & $r_{rel} (\SI{}{\degree})$ & $t_{rel} (\SI{}{\percent})$ & $r_{rel} (\SI{}{\degree})$ & $t_{rel} (\SI{}{\percent})$ & $r_{rel} (\SI{}{\degree})$\\ 
\hline

00 & 14.27 & 5.29 & 4.14 & 1.92 & 65.27 & 6.23 & 18.24 & 2.69 & 25.29 & 7.37 & 0.70 & 0.25\\
02 & 4.07 & 1.48 & 5.58 & 2.44 & 57.59 & 4.09 & 4.37 & 1.18 & X & X & 0.76 & 0.23\\
05 & 3.00 & 1.40 & 3.40 & 1.50 & 16.76 & 4.06 & 19.22 & 3.54 & 26.01 & 10.62 & 0.40 & 0.16\\
07 & 3.60 & 2.06 & 3.15 & 2.48 & 17.52 & 5.38 & 23.61 & 4.11 & 24.53 & 10.83 & 0.50 & 0.28\\
08 & 2.93 & 1.32 & 4.08 & 1.79 & 24.02 & 3.05 & 24.18 & 2.47 & 32.40 & 12.13 & 1.05 & 0.32\\
\hline
\end{tabular}
\end{center}
\end{table*}
\def\arraystretch{1.05}
\begin{table*}[t]
\caption{Comparisons to \ac{VIO} approaches on KITTI Odometry sequence 10. Results reproduced from box plots in \cite{clark2017} and medians, first quartiles, and third quartiles were estimated from figures. \cite{clark2017} only reported errors on distances of \SIlist[list-units=single]{100;200;300;400;500}{\meter} from KITTI Odometry sequence 10 so we only report identical results for VIOLearner in this table. Full results for VIOLearner on sequence 10 can be found in Tab. \ref{vio_0910_results}.}
\label{VIO_results}
\begin{center}
\begin{tabular}{l*{9}{c}}
\hline
 & \multicolumn{3}{c}{\textit{VIOLearner}} & \multicolumn{3}{c}{\textit{VINet}} & \multicolumn{3}{c}{\textit{EKF+VISO2}} \\
\cmidrule(lr){2-4} 
\cmidrule(lr){5-7} 
\cmidrule(lr){8-10} 
 
Length\\
\hline
& Med. & 1st Quar. & 3rd Quar. & Med. & 1st Quar. & 3rd Quar. & Med. & 1st Quar. & 3rd Quar. \\ 
\hline
100 & 1.87 & 1.25 & 2.3 & $\approx$ 0 & $\approx$ 0 & $\approx$ 2.18 & $\approx$ 2.7 & $\approx$ 0.54 & $\approx$ 9.2\\
200 & 3.57 & 2.9 & 4.16 & $\approx$ 2.5 & $\approx$ 1.01 & $\approx$ 5.43 & $\approx$ 11.9 & $\approx$ 4.89 & $\approx$ 32.6\\
300 & 5.78 & 5.19 & 6.39 & $\approx$ 6.0 & $\approx$ 3.26 & $\approx$ 17.9 & $\approx$ 26.6 & $\approx$ 9.23 & $\approx$ 58.1\\
400 & 8.32 & 6.87 & 9.57 & $\approx$ 10.3 & $\approx$ 5.43 & $\approx$ 39.6 & $\approx$ 40.7 & $\approx$ 13.0 & $\approx$ 83.6\\
500 & 12.33 & 9.49 & 13.69 & $\approx$ 16.8 & $\approx$ 8.6 & $\approx$ 70.1 & $\approx$ 57.0 & $\approx$ 19.5 & $\approx$ 98.9\\
\hline
\end{tabular}
\end{center}
\end{table*}
\def\arraystretch{1.05}
\begin{table*}[t]
\caption{Comparisons to \ac{VO} and \ac{VIO} approaches on KITTI sequences 09 and 10.  $t_{rel} (\SI{}{\percent})$ is the average translational error percentage on lengths \SI{100}{\meter} - \SI{800}{\meter}  and $r_{rel} (\SI{}{\degree})$ is the rotational error $(\SI{}{\degree}/\SI{100}{\meter})$ on lengths \SI{100}{\meter} - \SI{800}{\meter} calculated using the standard KITTI benchmark \cite{Geiger2013}.  }
\label{vio_0910_results}
\begin{center}
\begin{tabular}{l*{10}{c}}
\hline
 & \multicolumn{2}{c}{\textit{VIOLearner}} & \multicolumn{2}{c}{\textit{SFMLearner}} & \multicolumn{2}{c}{\textit{OKVIS}} & \multicolumn{2}{c}{\textit{ROVIO}} & \multicolumn{2}{c}{\textit{ORB-SLAM2}}\\
\cmidrule(lr){2-3} 
\cmidrule(lr){4-5} 
\cmidrule(lr){6-7} 
\cmidrule(lr){8-9} 
\cmidrule(lr){10-11} 
 
Seq \\
\hline
& $t_{rel} (\SI{}{\percent})$ & $r_{rel} (\SI{}{\degree})$ & $t_{rel} (\SI{}{\percent})$ & $r_{rel} (\SI{}{\degree})$ & $t_{rel} (\SI{}{\percent})$ & $r_{rel} (\SI{}{\degree})$ & $t_{rel} (\%)$ & $r_{rel} (\degree)$ & $t_{rel} (\SI{}{\percent})$ & $r_{rel} (\SI{}{\degree})$\\
\hline

09 & 1.51 & 0.90 & 21.63 & 3.56979 & 9.77 & 2.97 & 20.18 & 2.09 & 0.87 & 0.27 \\

10 & 2.04 & 1.37 & 20.54 & 10.93 & 17.30 & 2.82 & 20.04 & 2.24 & 0.60 & 0.27\\
\hline

\end{tabular}
\end{center}
\end{table*}

\begin{figure*}[t]
\label{kitti_trajectories}
\centering    
  \begin{subfigure}[KITTI 09]{0.49\textwidth}      
	\def\svgwidth{90mm}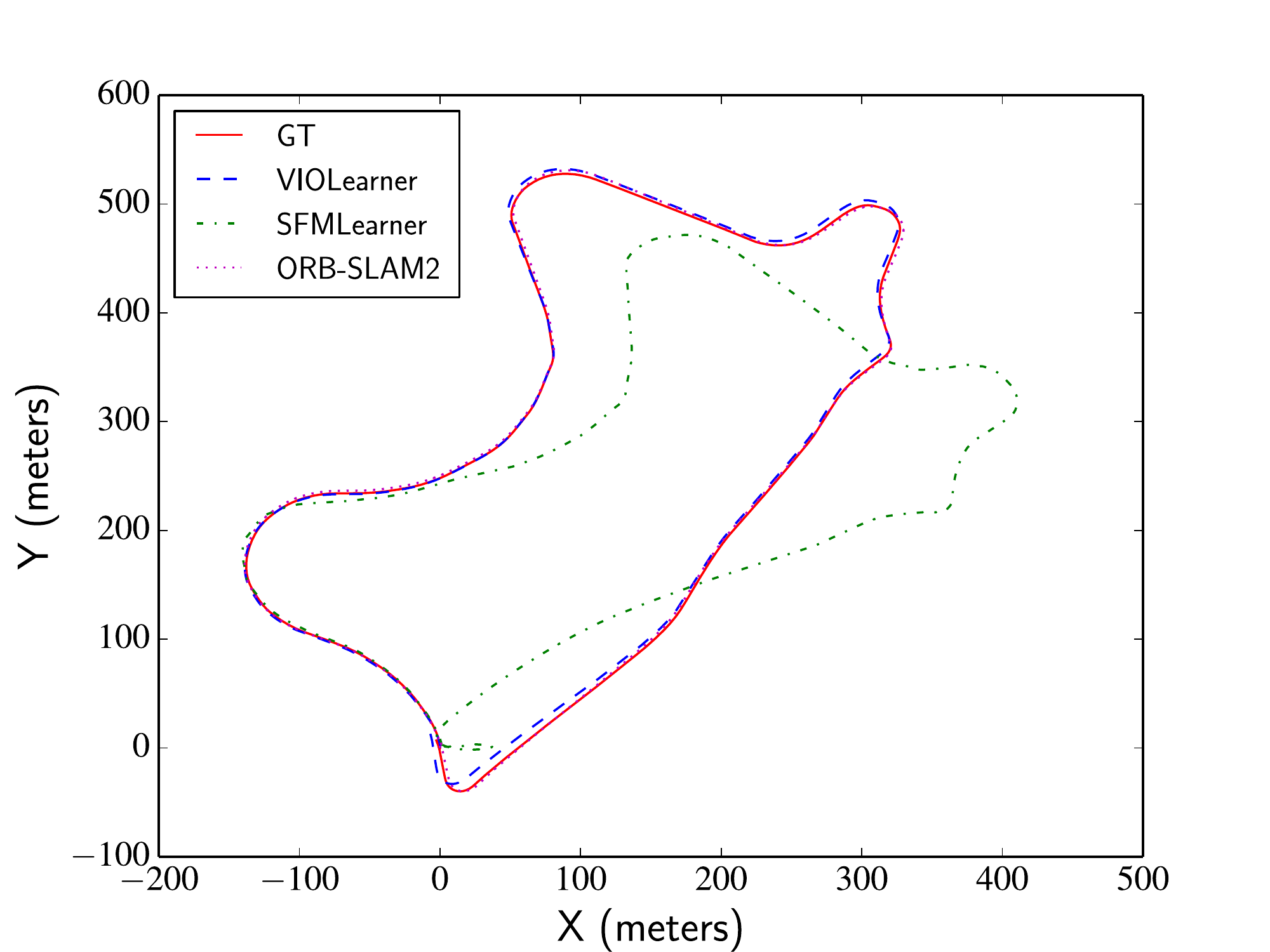	
 	\caption{KITTI sequence 09}
	\label{chapter2:entropy0} 	
  \end{subfigure}	
  \begin{subfigure}[No noise]{0.49\textwidth}      
	\def\svgwidth{90mm}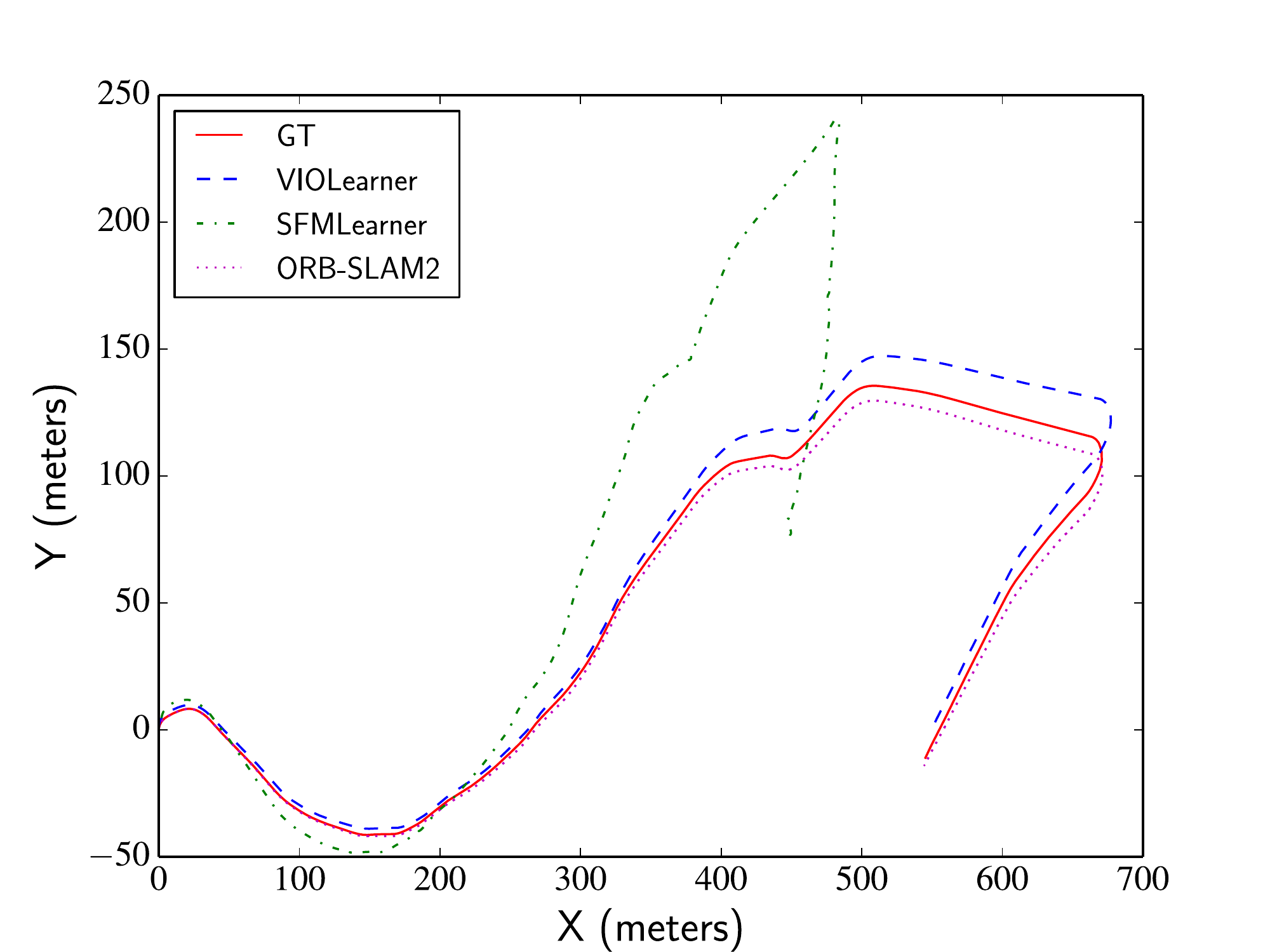	
 	\caption{KITTI sequence 10}
	\label{chapter2:entropy01} 	
  \end{subfigure}	
\caption{Trajectories on KITTI sequences 09 and 10 for VIOLearner, SFMLearner, ORB-SLAM2, and ground truth. }  
\end{figure*}

\begin{figure}[t]
	\centering
	\begin{subfigure}{0.49\columnwidth}
	\includegraphics[width=\columnwidth]{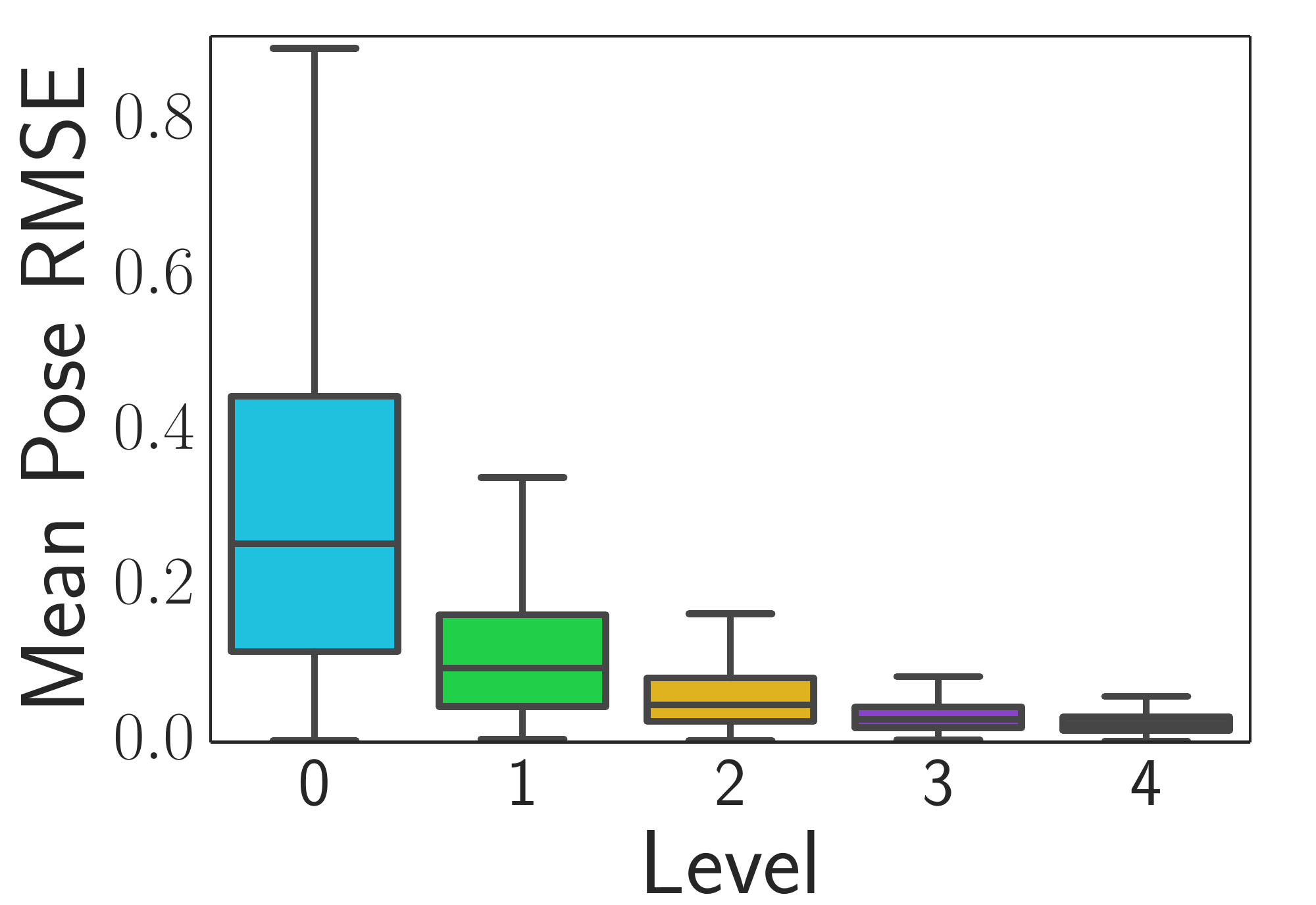}
	\label{fig:error_by_level_9}
	\caption{KITTI $09$}
	\end{subfigure}
	\begin{subfigure}{0.49\columnwidth}
	\includegraphics[width=\columnwidth]{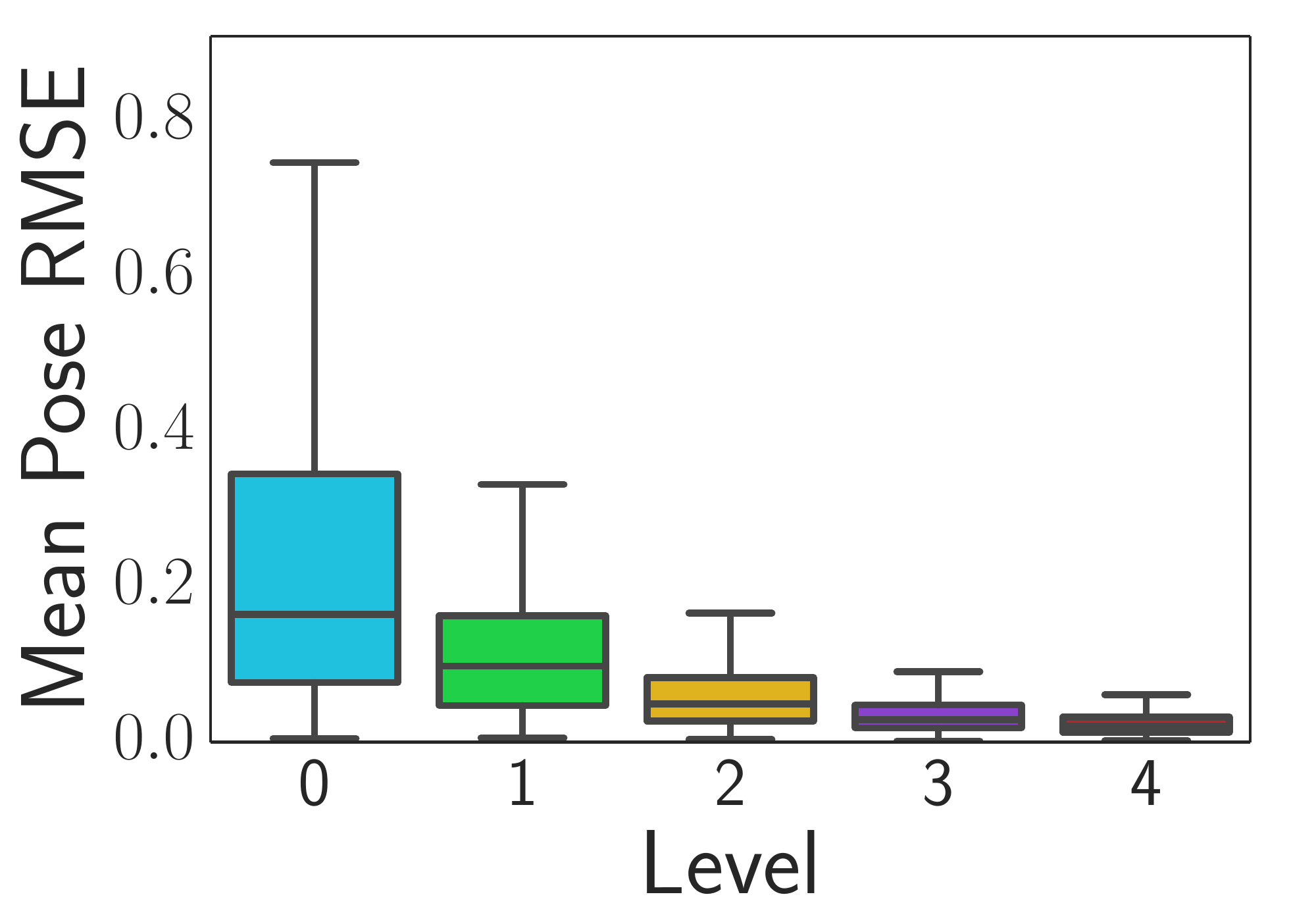}
	\label{fig:error_by_level_10}
	\caption{KITTI $10$}
	\end{subfigure}	
	\caption{Pose root mean squared error (RMSE) from poses computed by VIOLearner compared to ground truth after each level for KITTI sequences 09 and 10. We see a substantial decrease in error after repeated applications of our error correction module.} 
	\label{fig:error_by_level}
\end{figure}

%

\section{Results and Discussion} \label{results}

\subsection{Visual Odometry}

VIOLearner compared favorably to the \ac{VO} approaches listed above as seen in ÒTab. \ref{VO_results}Ò. It should be noted that the results in ÒTab. \ref{VO_results}Ò for VIOLearner, UnDeepVO, and SFMLearner are for networks that were tested on data on which they were also trained which is in accordance with the results presented in \cite{li2017}. We were thus unable to properly evaluate UnDeepVO against VIOLearner on test data that was not also used for training as such results were not provided for UnDeepVO nor is their model available online at the time of writing. With the exception of KITTI sequence 00, VIOLearner performed favorably against UnDeepVO.

We did however also evaluate VIOLearner more conventionally by training on sequences $00-08$ and testing on sequences 09 and 10 as was the case for \cite{zhou2017}. These results are shown in ÒTab. \ref{vio_0910_results}Ò. VIOLearner significantly outperformed SFMLearner on both KITTI sequences 09 and 10.

\subsection{Visual Inertial Odometry}

The authors of VINet \cite{clark2017} provide boxplots of their method's error compared to several state-of-the-art approaches for \SI{100}{\meter} - \SI{500}{\meter} on KITTI Odometry. We have extracted the median, first quartile, and third quartile from their results plots to the best of our ability and included them in ÒTab. \ref{VIO_results}Ò. For longer trajectories (\SIlist[list-units=single]{300;400;500}{\meter}), VIOLearner outperformed VINet on KITTI sequence 10.  It should again be noted that while VINet requires camera-\ac{IMU} extrinsic calibrations, our network is able to implicitly learn this transform from the data itself.

The authors of \cite{clark2017} reported OKVIS failing to run on the KITTI Odometry dataset and instead used a custom \ac{EKF} with VISO2 as a comparison to traditional \ac{SOA} \ac{VIO} approaches. However, we were able to successfully run KITTI Odometry on sequences 09 and 10 and have included the results in ÒTab. \ref{vio_0910_results}Ò. Additionally, we provide results from ROVIO. VIOLearner outperforms OKVIS and ROVIO on KITTI sequences 09 and 10. However, both OKVIS and ROVIO require tight synchronization between \ac{IMU} measurements and images which KITTI does not provide. This is most likely the reason for the poor performance of both approaches on KITTI. This also highlights a strength of VIOLearner in that it is able to compensate for loosely temporally synchronized sensors without explicitly estimating their temporal offsets. 

\subsection{Visual Simultaneous Localization and Mapping}

Additionally, we have included benchmark results from ORB-SLAM2. ORB-SLAM2 performs \ac{SLAM}, unlike our pure odometry-based solution, and is included as a example of \ac{SOA} localization to provide additional perspective on our results. VIOLearner was significantly outperformed by ORB-SLAM2. This was not a surprise as ORB-SLAM2 uses bundle adjustment, loop closure, maintains a \ac{SLAM} map, and generally uses far more data for each pose estimate compared to our VIOLearner.


\subsection{Online Error Correction}
Results in ÒFig. \ref{fig:error_by_level}Ò show the pose \acf{RMSE} between the $\hat{\theta}$ generated at each level and suggest that our online error correction mechanism is able to reduce pose error. It should however be noted that $\hat{\theta}$ for Levels 0 to 2 are computed using down-sampled images and thus their Jacobian inputs both have access to less information and use one less convolutional layer each. The extent to which this affects the plots in ÒFig. \ref{fig:error_by_level}Ò is as of yet not fully clear. However, Level 3 operates on full-size inputs and there is still a reduction in error between Level 3 and Level 4. While this reduction in the final layer can be partially attributed to the multi-hypothesis mechanism in Level 4, the mean \ac{RMSE} for Level 3 is $0.042$ while the mean \ac{RMSE} from each individual hypothesis pathway is $0.0375$, $0.0555$, $0.0262$, $0.0303$, $0.0332$ ($0.0261$ when the lowest error hypothesis is chosen) for KITTI 09 and $0.029$, $0.058$, $0.034$, and $0.037$  ($0.027$ when the lowest error hypothesis is chosen) for KITTI 10. The lower mean \ac{RMSE} from individual Level 4 hypotheses (with the exception of the second hypothesis pathway) suggests that the observed error reduction is indeed an effect of our online error correction mechanism rather than simply an artefact of image resolution.

\subsection{Multi-Hypothesis Error Reduction} \label{multi_hypoth_disc}
For the $\hat{\theta}$s generated by each of the four hypothesis pathways in the final level for KITTI sequence 09, the average variance of pose error between the four hypotheses was $2.7\mathrm{e}{-5}$ in the x-dimension, $3\mathrm{e}{-6}$ in the y-dimension, $4.69\mathrm{e}{-4}$ in the z-dimension, and $4.06\mathrm{e}{-4}$ was the Euclidean error between the computed translation and the true translation. The z-dimension shows $1-2$ orders of magnitude more variance compared to the x- and y- dimensions and is the main contributor to hypothesis variance. For the camera frame in KITTI, the z-direction corresponds to the forward direction which is the predominant direction of motion in KITTI and also where we would expect to see the largest influence from uncertainty in initial velocities. These results are consistent with the network learning to model this uncertainty in initial velocity as intended.
 
\section{Conclusion} \label{conclusion}
In this work, we have presented our VIOLearner architecture and demonstrated competitive performance against \ac{SOA} odometry approaches. VIOLearner's novel multi-step trajectory estimation via convolutional processing of Jacobians at multiple spatial scales yields a trajectory estimator that learns how to correct errors online. 

The main contributions of VIOLearner are its unsupervised learning of scaled trajectory, online error correction based on the use of intermediate gradients, and ability to combine uncalibrated, loosely temporally synchronized,  multi-modal data from different reference frames into improved estimates of odometry.





%
%

%


\bibliographystyle{IEEEtran}   			
\bibliography{IEEEabrv,vio}

\end{document}